# Content associations with answer selection and engagement in supply chain community question answering

Hisatoshi Naganawa[1*]

*Correspondence:
Hisatoshi Naganawa
2117053z@gsuite.kobe-u.ac.jp
[1]International Maritime Research Center, Kobe University, Kobe, Japan

**Abstract**
Community question-answering (cQA) platforms are important ecosystems for knowledge sharing, yet research on what makes an answer "best" or a question engaging has often focused on either individual lexical features or broad topics. This exploratory study examines both levels in Yahoo! Chiebukuro, asking which nouns and semantic noun clusters are associated with best-answer selection and question engagement. We analyze a large corpus of questions and answer-tier posts collected under the keyword "logistics," using an offline machine-translation pipeline with manually validated noun-level fidelity. Best-answer selection is modeled with conditional logistic regression within question threads, controlling for answer length, order, and question-answer noun overlap. Engagement is modeled in two stages: whether a question receives any response and, conditional on receiving one, how many responses it attracts. Operational logistics nouns such as "distribution," "warehouse," and "center" are associated with best-answer selection, whereas broader civic and infrastructure nouns such as "car," "tax," and "shinkansen" are associated with higher engagement. At the cluster level, most semantic noun clusters are associated with best-answer selection, while the clusters associated with response likelihood and response volume are largely distinct. These findings reveal differentiated content-outcome profiles across answer and question levels and provide a domain-wide descriptive map of engagement and answer selection in logistics-related cQA discourse.

## 1 Introduction

On a community question-answering (cQA) platform, two structural outcomes indicate whether a post is seen as valuable: whether an answer is selected as "best," and whether a question generates sustained back-and-forth. Both outcomes are, in principle, related to content — an answer is chosen partly because of what it says, a question attracts replies partly because of what it asks — yet studies of platform discourse often examine content either at the level of broad topics (what the community talks about) or at the level of individual lexical features, rather than connecting the two, and rarely ask whether a content-outcome association survives once an obvious confound like post length is taken into account. This paper is an exploratory empirical study that works across both levels, using Yahoo! Chiebukuro discourse on logistics as a case. Because the analysis is restricted to nouns extracted from a machine translation, it characterizes the topical lexical content of posts — which referential nouns and semantic clusters they contain — rather than wording in the broader linguistic sense (style, politeness, hedging, sentiment).

Logistics is a useful case for this kind of content-outcome analysis. It is a professional, operational domain where questions range from narrow transactional queries (tracking a specific parcel, comparing shipping services) to broad, debatable civic and infrastructure topics (road congestion, transportation policy, retail supply chains), giving the corpus real variation in both the concreteness of its content and the structural fate of its posts.

We ask three questions. First, which nouns are associated with an answer being selected as "best" rather than passed over on the same question? Second, which nouns are associated with a question attracting many answers and replies rather than few or none? Third, do these word-level patterns aggregate into coherent semantic clusters, and are those clusters still associated with the two outcomes once the length of the post — and, for the question-level models, its posting year — is statistically controlled? Throughout, the study is observational: it reports conditional associations between content and outcomes, not causal effects, and the answer-level and question-level analyses are associations measured on different units of text.

The remainder of this paper is organized as follows. Section 2 reviews related work on platform affordances, online knowledge sharing, answer selection, and social media discourse in logistics contexts. Section 3 describes the data collection, translation, and preprocessing pipeline, and the keyness and cluster-regression methods used to analyze it. Section 4 presents the results: the noun vocabulary associated with best-selected answers and high-engagement questions, the semantic cluster structure recovered from the corpus, and which clusters are associated with each outcome. Section 5 discusses these results together with the study's limitations and avenues for future research, and Sect. 6 concludes.

## 2 Related work

### 2.1 Platform affordances and structural constraints

Platform architecture shapes how users interact and produce content [5]. The concept of "affordances" — coined by Gibson [5, 17] and adapted to technology by Norman [24] and Treem and Leonardi [33] — refers to the actionable properties between an actor and an object; on social platforms these affordances bound the space of possible interaction modes [4, 33]. Community question-answering platforms such as Yahoo! Chiebukuro, Quora, and Stack Exchange are explicitly built around asynchronous, reciprocal information-seeking [1, 29]; thread loops, best-answer selection, and reputation or incentive systems shape the quality and structure of the resulting exchange [7, 31].

### 2.2 Online knowledge sharing and interaction modes

Online knowledge sharing depends on participants' willingness to externalize tacit knowledge into explicit digital form [23, 34]. In cQA environments this exchange is largely goal-oriented and problem-driven [2, 10]: participants are motivated to contribute operational narratives and practical knowledge by reputation, reciprocity, and self-efficacy [6, 15]. Studies of Yahoo! Chiebukuro specifically document it as a venue where users request and disseminate everyday, often highly personal, domain-specific concerns and knowledge [30]. What this literature has not examined is whether the specific lexical content associated with an answer's "best" status, or with a question's sustained attention, can be characterized systematically across a large corpus, at a level of detail finer than topic. A substantial body of work examines answer selection and quality directly, using lexical, semantic, user, and interaction features to predict which answers are chosen or rated helpful [3, 14]. More recent studies emphasize the relationship between an answer and its question: perceived helpfulness rises when an answer's language attributes, including concreteness, are congruent with the question's [26], and thematic consistency between answer and question is positively related to helpfulness while sentiment consistency is not [16]. On the question side, how a question is crafted — including strategies affecting its concreteness — is correlated with both audience engagement and answer quality [8]. It is important to note that an asker-selected "best answer" reflects the asker's subjective judgment rather than objectively verified correctness [12, 29]. Accordingly, this study treats best-answer selection as an indicator of perceived value within the platform, rather than as a measure of answer correctness. Relative to this predictive and experimental literature, the present study takes a descriptive approach. Rather than seeking to maximize predictive performance, we examine a domain-specific corpus to identify which individual nouns and semantic clusters are associated with each outcome.

### 2.3 Social media in logistics and supply chain discourse

Social media research has extensively covered consumer behavior, political discourse, and entertainment, but its application to professional discourse in logistics and supply chain management remains comparatively underdeveloped [11, 13]. Logistics knowledge communication has traditionally been studied through formal organizational channels and B2B networks [19, 27]; recent supply-chain disruptions have driven practitioners toward informal digital spaces for real-time insight and peer validation [28], and recent reviews explicitly call for more social-media-based analysis in supply-chain and logistics contexts [25, 32]. A related recent strand applies NLP and large-language-model methods to logistics-relevant social media text from a different platform, YouTube: one study extracts logistics challenges from video subtitles and comments and characterizes them into recurring concern areas — time management, driver welfare, technological investment, policy transparency [22] — and a companion study uses the same YouTube corpus to generate policy proposals via graph-based retrieval-augmented generation [21]. Both target aggregate discourse characterization or policy generation rather than the fine-grained, word- and cluster-level question this paper addresses on a different platform: which specific content is associated with a post's structural fate.

### 2.4 Research gap

Prior work establishes that (a) platform affordances structure interaction, (b) cQA discourse is characteristically problem-driven, and (c) answer selection and question engagement have each been modeled with rich feature sets — but logistics-specific discourse on social media remains comparatively understudied, and few studies map an entire domain corpus's noun vocabulary and its semantic-cluster structure against both outcomes at once. This paper addresses that gap as an exploratory empirical study: it combines a keyness scatter method that reveals the full vocabulary associated with each outcome with length-controlled cluster-level regressions, using established methods (keyness statistics, Word2Vec, HDBSCAN, conditional and count regression) whose combination is offered as a practical workflow rather than a new analytical framework.

## 3 Methods

### 3.1 Data collection

Question posts, best/other answers, nested replies, and asker acknowledgements were collected from Yahoo! Chiebukuro via a Selenium-based scraper using the search keyword "物流" (logistics), spanning April 2004 to July 2026. To avoid the platform's cap on the number of retrievable search results, the search was partitioned into one-day posting-date windows, each paginated to exhaustion; question URLs were de-duplicated before thread pages were fetched, and each thread page was parsed into its question, answers, nested replies, and thanks comments. Posts deleted or inaccessible at crawl time are simply absent from the corpus, and questions still open at crawl time are retained but excluded from the best-answer analyses. The corpus comprises 40,871 questions and 78,641 answer-tier posts, tagged by structural role: best answer ($n = 24{,}512$), other answer ($n = 36{,}314$), reply ($n = 7{,}307$), and thanks ($n = 10{,}508$). Of the 40,871 questions, 24,767 (61%) received at least one answer-tier post. Temporally, the corpus is not front-loaded despite the 22-year span: yearly question volumes range from 80 (2004) to 3,907 (2011) and remain between roughly 1,500 and 2,400 per year from 2013 onward, with a median question year of 2016 and 36% of questions posted in 2020 or later (Appendix Figure A1). Because posting spans two decades, all question-level regressions below control for posting year Fig. 1.

### 3.2 English translation

Every question and answer-tier text was machine-translated from Japanese to English with a local, offline OPUS-MT Japanese–English model (Argos Translate package translate-ja_en v1.1, converted to CTranslate2), run directly through its SentencePiece tokenizer and the CTranslate2 Translator [9]. Long posts are split into sentence-boundary chunks (fragments over 200 characters further split), translated, and reassembled — no content discarded. Translation quality is below a commercial system's, but since the analysis operates on noun-level aggregate statistics rather than sentence meaning, this is expected to act mainly as noise rather than a directional bias, provided errors are uncorrelated with outcome.

To verify whether the translations appropriately retained the nouns under analysis, the author—a native Japanese speaker—manually reviewed 50 examples (25 questions and 25 answers) selected evenly from the short, medium, and long length groups. In 34 of the 50 pairs (68%), all key content nouns were retained, and in the remaining cases, the

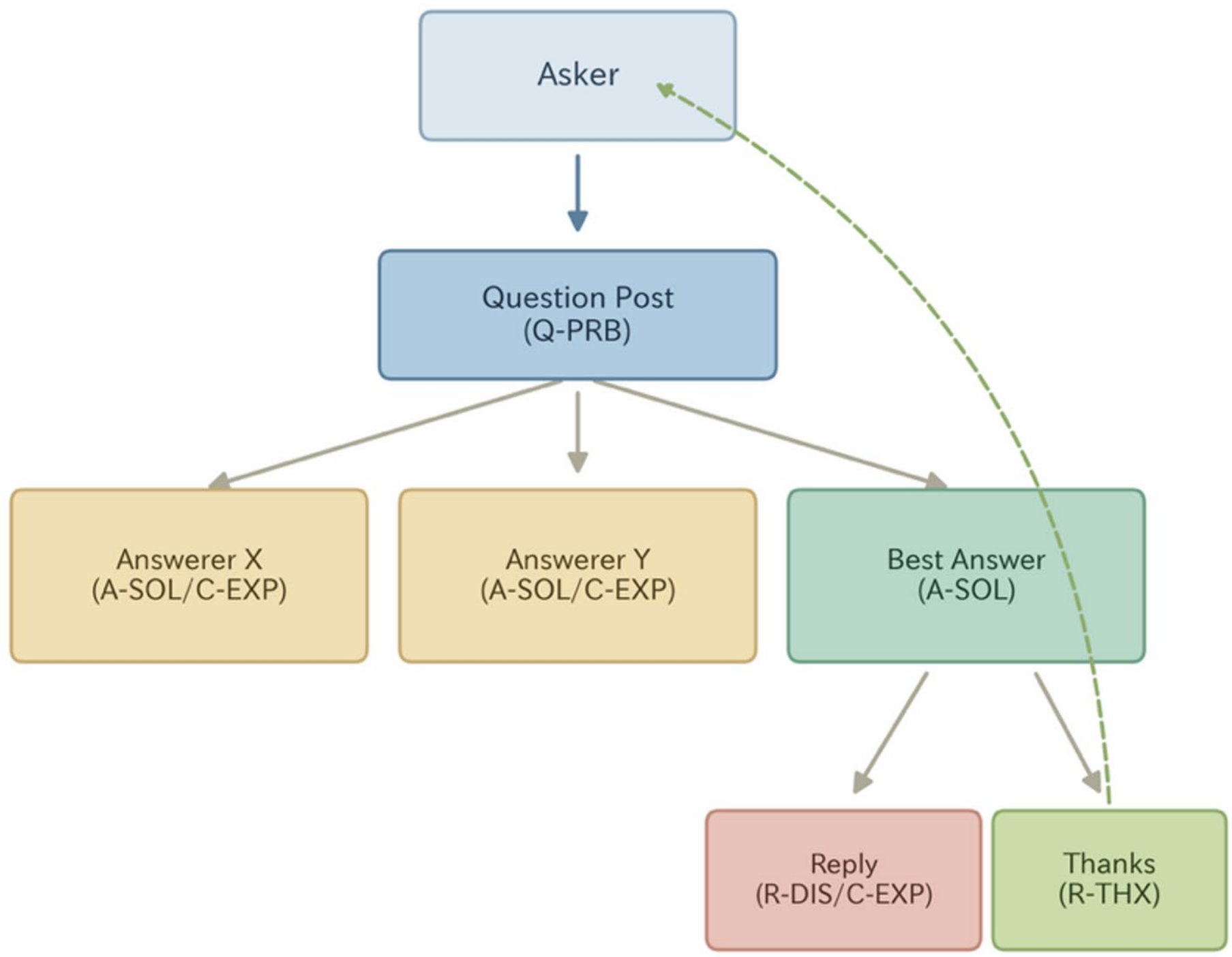

**Fig. 1** cQA Thread Structure

majority of content nouns were retained. Errors were primarily limited to proper nouns and rare words, while key topic-related nouns were consistently retained. Furthermore, no translation errors that altered the topic itself were identified. Based on the above, we concluded that the machine translation system used in this study possesses sufficient accuracy for content analysis at the noun level. Furthermore, certain translation and part-of-speech tagging artifacts identified during verification—specifically fragments of contractions such as “don” and “isn,” invalid word fragments such as “gnation,” and formulaic polite expressions such as “please” and “thank” that were incorrectly tagged as nouns—were excluded as stop words.

### 3.3 Preprocessing and noun extraction

All analysis operates on the English translation. Tokenization and POS-tagging used NLTK 3.9.2, restricted to nouns (NN*), lemmatized (WordNet), and lowercased. Retained tokens are either alphabetic tokens of length > 2 or numeric tokens of ≥ 4 digits (keeping year references such as “2024” while dropping tracking-number fragments); a standard English stopword list plus a hand-curated list removes generic non-content nouns (e.g., “thing,” “way”), contraction fragments and translation artifacts, and formulaic politeness tokens. Restricting to nouns isolates topical content; verbs, adjectives, and pragmatic/stylistic features are outside this paper’s scope.

### 3.4 Content-driven selection and engagement analysis

Two descriptive comparisons examine which nouns are associated with a post’s structural fate, using the informative-Dirichlet-prior log-odds z-score of Monroe, Colaresi, and Quinn, which uses the pooled corpus as its own background prior and penalizes rare words [20]. Word counts are document-level, so a single post's repetition cannot dominate the statistic; words with fewer than 15 combined occurrences were discarded.

The best-answer comparison restricts to threads with ≥ 2 competing answers and an explicit best answer (15,104 threads, 35,980 non-best answers), comparing the chosen best answers against the passed-over answers. Because this pooled comparison ignores the matched thread structure and weights questions by their number of answers, the headline nouns are additionally validated with a within-thread permutation test: the best/other labels are shuffled 500 times within each thread, and each reported noun's observed $|z|$ is compared against its permutation distribution. The engagement comparison splits questions with ≥ 1 answer-tier post into high (≥ 75th percentile: ≥ 4 responses; $n = 6{,}747$) and low (≤ 25th percentile: ≤ 1 response; $n = 9{,}602$) groups by total answers plus replies, comparing question-text word usage; the percentile split is used only for this descriptive lexical contrast, and a tertile split yields closely overlapping top-noun lists.

### 3.5 Semantic noun clustering

Word-level keyness cannot show whether related nouns form a coherent group. A continuous bag-of-words Word2Vec model (gensim 4.4.0, single-threaded for bit-reproducibility) was trained on the noun-tokenized corpus, and the embeddings of the 800 most frequent nouns were L2-normalized, PCA-reduced, and clustered with HDBSCAN [18]. Parameters were fixed by a pre-stated rule over a grid of PCA dimensionality, minimum cluster size, minimum samples, and cluster-selection epsilon (Table 3): among settings yielding 6 to 15 clusters whose largest holds at most 50% of classified nouns, take the one maximizing frequency-weighted coverage. Unassigned words ("noise") stay unclassified. The chosen solution (12 clusters, 50.3% of noun-token frequency) is validated in Table 8 for coherence, parameter stability, and sensitivity to dropping reply and thanks posts, with each cluster's similarity to the "logistics" vector as an interpretive aid. Membership is in Table 11. For visualization only (Fig. 4), the same vectors are projected with UMAP.

### 3.6 Cluster-level regressions

To separate a cluster's association with an outcome from simply writing more, each post was reduced to cluster-presence indicators plus log(1 + character length); a robustness check replaces indicators with log(1 + cluster token count). Best-answer selection is modeled with a conditional logistic regression stratified by question, comparing each selected answer against its own competitors and thereby absorbing all question-level characteristics. Controls are log length, chronological order within the thread, and question-answer noun overlap (Jaccard similarity). The originally submitted pooled logistic model with question-clustered standard errors is retained as a robustness check. Question engagement is modeled in two parts so that the 16,104 questions (39%) without responses are not excluded: a logistic regression for receiving any response, and a negative binomial for the number of responses among answered questions (Poisson is overdispersed, Pearson $\chi^2/\mathrm{df} \approx 1.45$). Both use the 40,836 questions with a usable timestamp and non-empty text, of which 24,756 are answered, and control for log length and centered posting year. Robustness checks separate answers from replies and refit on the 2004–2014 and 2015–2026 subperiods.

All models report 95% CIs, exponentiated effect sizes, and Benjamini-Hochberg FDR-corrected p-values across the twelve cluster terms within each model. Results are conditional associations, not causal effects: answerer reputation, question difficulty, and platform recommendation are unmeasured.

## 4 Results

### 4.1 Nouns associated with best-answer selection

The keyness scatter in Fig. 2 shows a clear pattern: answers chosen as best are distinguished by concrete nouns naming logistics operations, while passed-over answers are distinguished by generic, often off-topic nouns. These are answer-level lexical associations, not evidence that any particular word causes selection.

Best-selected answers are characterized overwhelmingly by "logistics" itself (the single strongest noun in either direction by a wide margin, z = 38.1), followed by "distribution" (z = 7.7), "center" (z = 6.2), "warehouse" (z = 5.8), "base" (z = 5.4), "network" (z = 4.3), and "manufacturing" (z = 4.1). Passed-over answers are characterized by "people" (z = – 9.8), "job" (z = – 7.6), "car" (z = – 7.0), "reference" (z = – 7.0), "person" (z = – 6.9), "company" (z = – 5.6), and "party" (z = – 5.0). Inspection of the Japanese originals shows that "reference" derives mainly from the formulaic 参考 family — "参考までに" (for your reference) and the platform's 「参考にした回答」 quoted-answer footer — so it marks deferential or secondhand answers rather than literal citation of external sources. Because this pooled comparison weights threads by their number of answers, the reported nouns were re-tested with the within-thread permutation test. For all fourteen nouns the observed z exceeded every one of the 500 within-thread relabelings (permutation $p$ =.002, the resolution floor at 500 permutations).

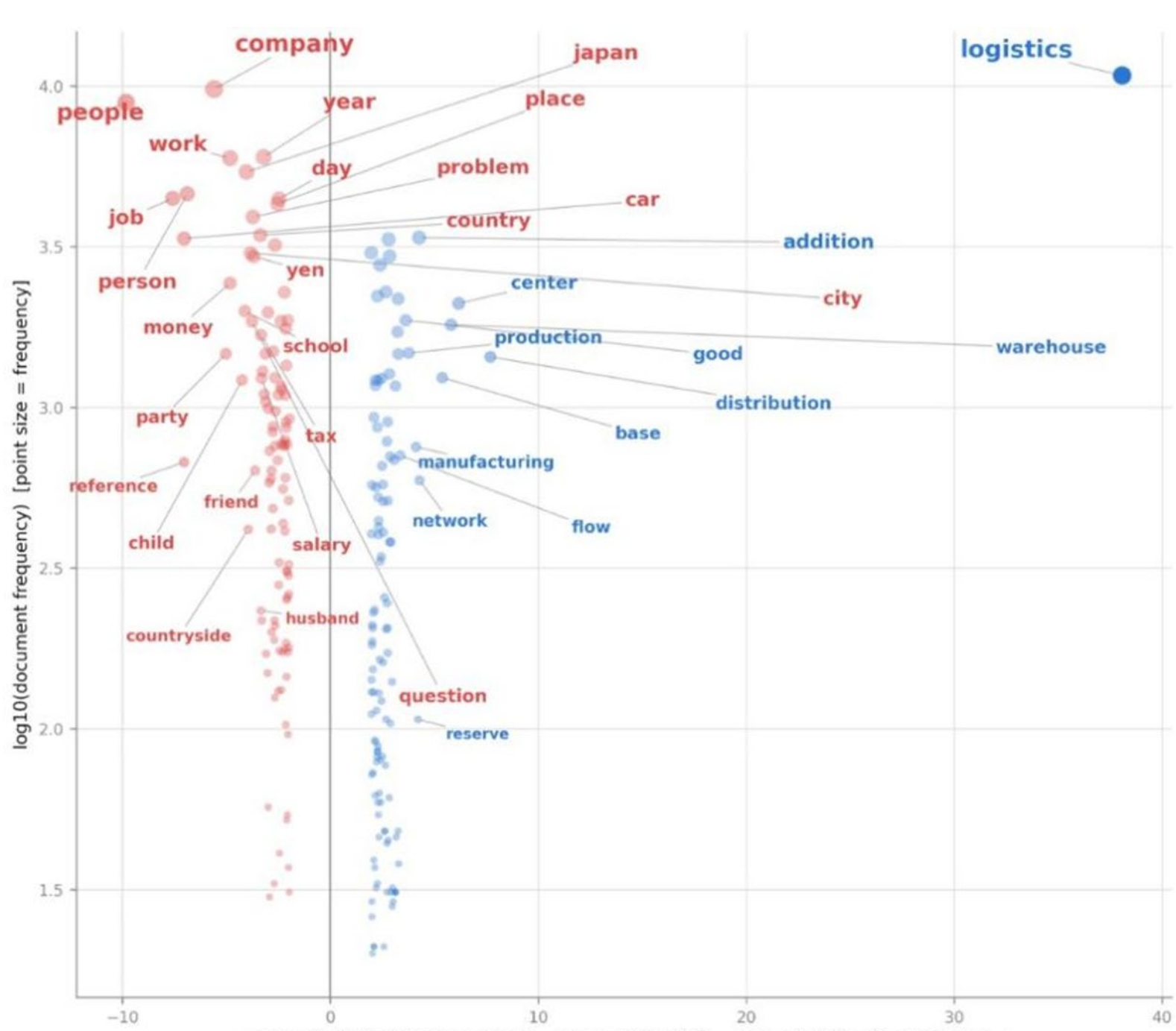

**Fig. 2** Lexical signature of best answers

### 4.2 Nouns associated with question engagement

The keyness scatter in Fig. 3 shows that high-engagement questions (many answers and replies) and low-engagement questions (typically answered once, if at all) are separated primarily by scope. These are question-level lexical associations.

High-engagement questions cluster around broad, debatable, civic and infrastructure nouns: "car" (z = 6.5), "war" (z = 4.9), "tax" (z = 4.8), "party" (z = 4.7), "shinkansen" (z = 4.3), "road" (z = 4.2), "expressway" (z = 4.1), "child" (z = 4.1), and "highway" (z = 4.1). Low-engagement questions, conversely, are dominated by narrow, transactional nouns: "logistics" itself is the strongest single low-engagement noun (z = – 9.7, the corpus's own search keyword), alongside "product" (z = – 6.6), "Taobao" (z = – 6.5, a cross-border e-commerce marketplace), "company" (z = – 6.3), "warehouse" (z = – 6.2), and "center" (z = – 6.1). Inspection of the Japanese originals of Taobao-related questions confirms the interpretation: they overwhelmingly concern individual cross-border purchases — stalled parcel tracking, customs holds, forwarding agents, refund requests — that is, single-answer transactional queries rather than discussion starters. The top-noun lists are robust to the split threshold: replacing the 75th/25th percentile split with a tertile split leaves 18 of the top 20 high-engagement nouns and 16 of the top 20 low-engagement nouns unchanged. Notably, "logistics," "center," "warehouse," and "distribution" — nouns associated with a best answer are associated with a low-engagement question here. Since the two comparisons are computed on different units (answers vs. questions), this is not a within-post trade-off; rather, it suggests that the same operational vocabulary plays different roles depending on where it appears.

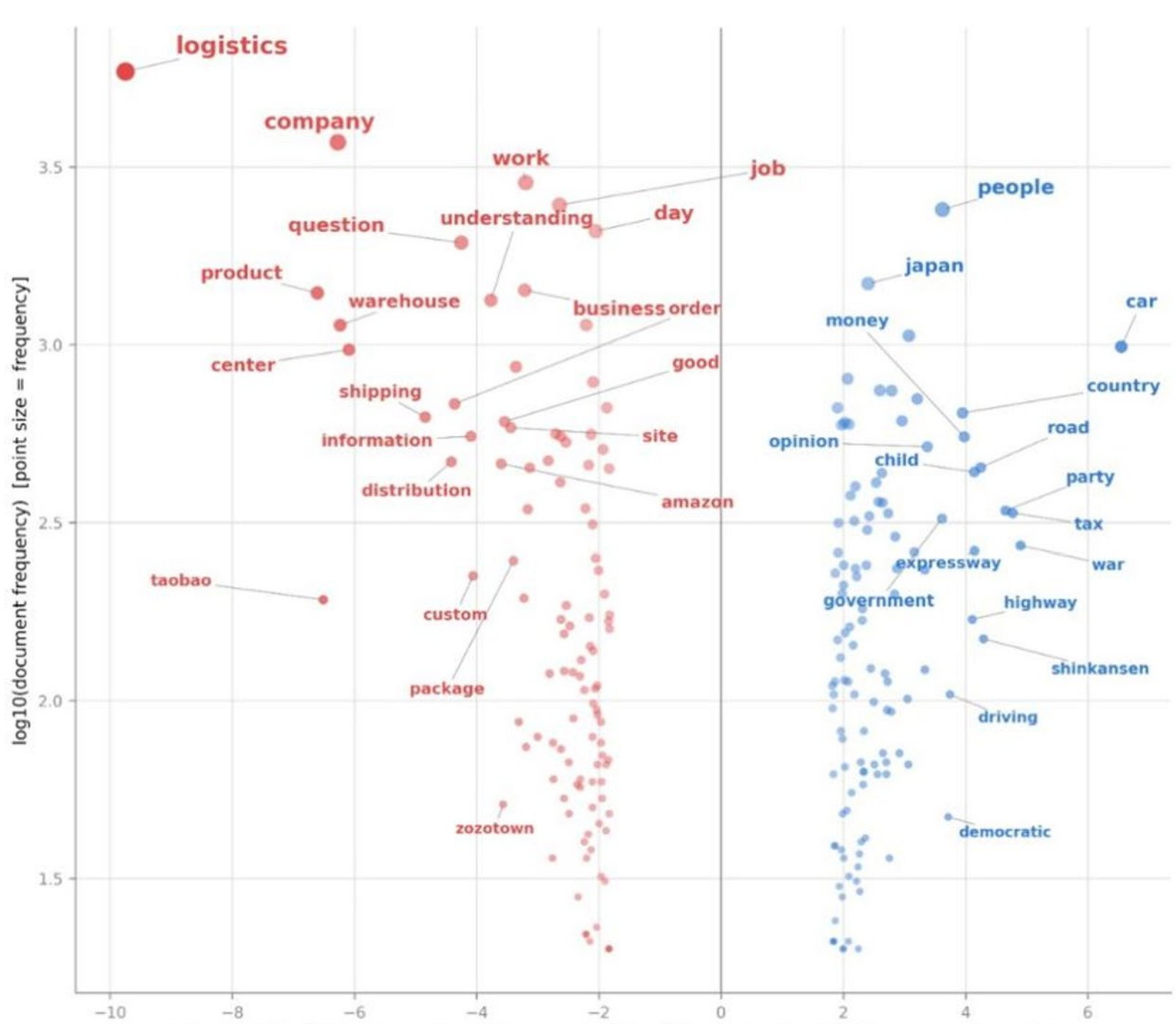

**Fig. 3** Question nouns and engagement

### 4.3 A semantic cluster map of the noun vocabulary

Applying the selection rule of § 3.5 to the 800 most frequent nouns yields twelve semantic clusters covering 302 nouns and 50.3% of noun-token frequency among the candidates. The solution is markedly more balanced than an unconstrained coverage-maximizing choice would be: the largest cluster holds 40 nouns (13% of classified nouns, 14% of total noun-token frequency), and no cluster mixes obviously unrelated registers Table 1.

Two caveats apply to reading these as substantive constructs. Cluster 12 (calendar and dates) is best understood as a temporal lexical artifact rather than a content theme, and Cluster 5 mixes job-seeking vocabulary with question-answering framing terms ("question," "answer," "reply"); both are retained because the selection rule was fixed in advance, but neither is interpreted substantively below. The 498 unclassified nouns (49.7% of noun-token frequency) are treated as background rather than forced into clusters — including "logistics" itself, the corpus's own search keyword and second most frequent noun, which co-occurs with essentially everything in a keyword-filtered corpus and therefore has no locally dense neighborhood Fig. 4.

Appendix Table 8 reports four validation checks (§ 3.5). All clusters show higher mean pairwise cosine similarity than random noun sets of the same size drawn from the 800 candidates, indicating that the clusters are not arbitrary partitions. Across eight neighboring parameter settings, the adjusted Rand index (ARI) averages 0.82, with six settings reproducing the reported partition almost exactly (ARI $\geq$ 0.97). The main sensitivity is to PCA dimensionality rather than HDBSCAN parameters: reducing it to 18 produces three coarse clusters (ARI = 0.03), whereas increasing it to 25 produces ten partially aligned clusters (ARI = 0.55). Removing reply and thanks posts has little effect on the semantic space, although it affects the clustering output. These posts account for only 3.3% of noun tokens, and 85% of the ten nearest neighbours for each classified noun are preserved (median 90%), compared with 2.6% under random pairing. Cluster coherence also remains essentially unchanged. However, HDBSCAN produces only two clusters when rerun on the reduced corpus. Together with the parameter-stability results, this suggests that the brittleness lies in the density-based clustering step rather than in the underlying semantic space. Thus, the semantic neighbourhoods are robust to corpus composition, whereas the exact twelve-cluster partition depends on the embedding and parameter settings. We therefore treat the reported partition as one explicit and reproducible solution rather than a uniquely determined structure.

**Table 1** Semantic Noun Cluster Definitions

| Cluster | Short name | Representative nouns |
|---|---|---|
| 1 | Work and employment | company, work, year, day, business, office, employee, department |
| 2 | Delivery and ordering | product, delivery, order, shipping, distribution, mail, luggage |
| 3 | Places and regions | city, area, tokyo, prefecture, osaka, region, port, hokkaido |
| 4 | Nations and geopolitics | japan, country, power, china, state, war, economy, korea |
| 5 | Job seeking and Q&A framing | job, question, understanding, experience, interview, answer |
| 6 | Road and rail transport | car, road, line, station, route, traffic, railway, expressway |
| 7 | Cost, tax and government | cost, tax, government, amount, party, rate, debt, consumption |
| 8 | Opinion and interpretation | person, word, story, feeling, idea, opinion, questioner, meaning |
| 9 | Education | school, university, student, test, class, english, teacher |
| 10 | Family and workplace relations | woman, family, parent, friend, workplace, boss, mother, husband |
| 11 | Infrastructure and strategy | support, development, resource, security, infrastructure, network |
| 12 | Calendar and dates | end, middle, march, beginning, april, june, december, january |

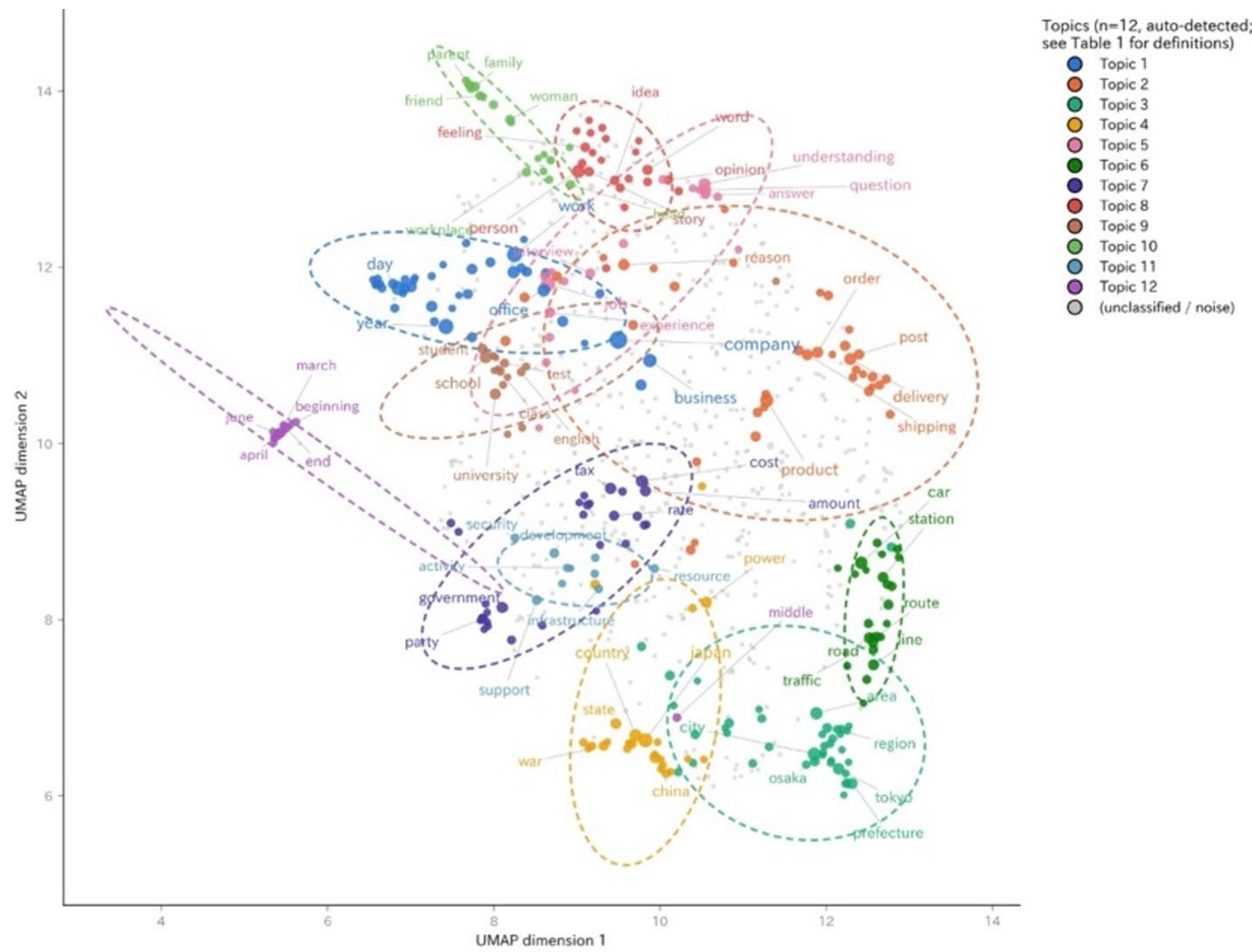

**Fig. 4** Semantic Noun Cluster Map (UMAP projection)

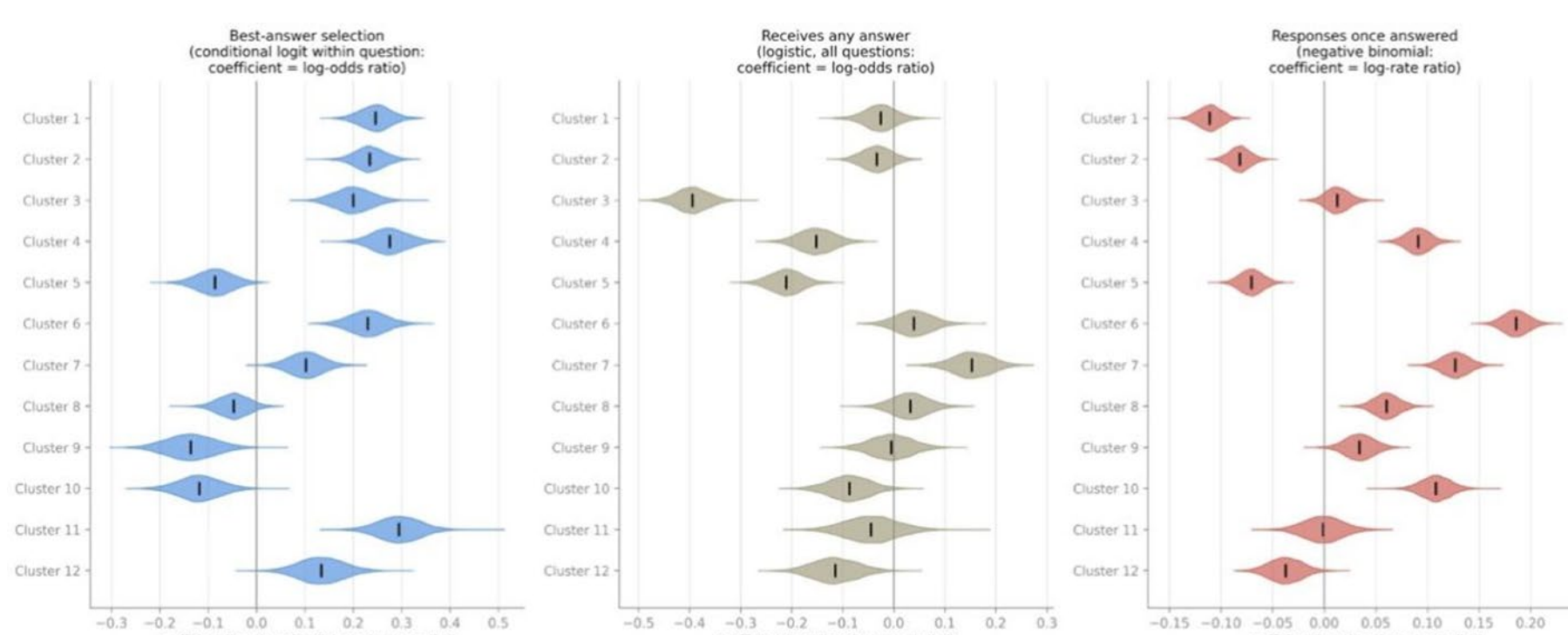

**Fig. 5** Cluster-level associations with best-answer selection, response likelihood, and response volume

### 4.4 Semantic clusters associated with best-answer selection and engagement, net of length

Figure 5 reports the three models: a conditional logit for best-answer selection within threads, a logistic model for whether a question is answered at all, and a negative binomial for how many responses an answered question receives. Full coefficients, standard errors, confidence intervals, effect sizes, and raw and FDR-corrected p-values are in Appendix Table 4.

For best-answer selection, eleven of twelve clusters are associated with selection at p_FDR < 0.05; only Cluster 8 (opinion and interpretation) is null. The largest positive associations are Cluster 11 (infrastructure and strategy, OR = 1.34, 95% CI [1.23, 1.46]), Cluster 4 (nations and geopolitics, 1.32 [1.23, 1.42]), Cluster 1 (work and employment, 1.28 [1.20, 1.36]), Cluster 2 (delivery and ordering, 1.26 [1.19, 1.34]), and Cluster 6 (road and rail transport, 1.26 [1.17, 1.36]). Negative associations are confined to personal or meta-discursive vocabulary: Clusters 9 (education, 0.87 [0.79, 0.96]), 10 (family and workplace relations, 0.89 [0.82, 0.97]), and 5 (job seeking and Q&A framing, 0.92 [0.85, 0.98]). The controls are informative in their own right. Answer length dominates (OR = 2.59 [2.48, 2.71] per unit of log length), later answers are somewhat favoured (1.14 [1.12, 1.16] per rank), and question–answer noun overlap is strongly negative (0.13 [0.09, 0.18]): answers that merely restate the question's vocabulary are passed over.

Whether a question is answered at all turns on largely different clusters, which is the payoff of separating the two parts. Lower odds attach to Cluster 3 (places and regions, OR = 0.67 [0.63, 0.72]), Cluster 5 (0.81 [0.76, 0.86]), Cluster 4 (0.86 [0.80, 0.92]), and Cluster 12 (calendar and dates, 0.89 [0.82, 0.97], not interpreted substantively); only Cluster 7 (cost, tax and government) raises them (1.17 [1.08, 1.25]). Seven clusters, including work and employment and delivery and ordering, are null. Question length is not significant (OR = 0.96, $p$ =.075). Posting year is significant (OR = 1.38 per year), but we read this as an archival artifact, since older threads have had longer to lose answers to deletion, and retain the term as a control without interpreting it.

Once a question is answered, the ordering changes again. Cluster 6 has the largest positive association on any outcome here (IRR = 1.20 [1.18, 1.23]), followed by Clusters 7 (1.14 [1.11, 1.16]), 10 (1.11 [1.08, 1.14]), and 4 (1.10 [1.07, 1.12]). The two clusters closest to routine operational content are negative: Cluster 1 (0.90 [0.88, 0.91]) and Cluster 2 (0.92 [0.90, 0.94]). The dispersion parameter is $\alpha = 0.13$, confirming overdispersion relative to Poisson.

Four robustness checks appear in Appendix Table 4. The earlier pooled logit agrees in sign with the conditional logit on 11 of 12 clusters, so stratification changes magnitudes rather than directions. Replacing binary indicators with log(1 + cluster token count) reproduces all 12 signs. Modelling answers and replies separately reproduces 12 and 11 signs, so the combined measure hides no opposing dynamics. Refitting the count model on 2004–2014 ($n$ = 3,225) and 2015–2026 ($n$ = 21,531) reproduces 9 and 11 signs; the earlier, noisier subperiod's three sign changes are not read as a temporal shift.

Read together, no cluster behaves the same way across all three outcomes, and the clusters that win selection within a thread are not those that attract volume. Cluster 7 is the only one positive on all three. Cluster 6 is positive for selection and volume but null for receiving any answer. Cluster 3 is positive for selection yet most negative for receiving any answer. Clusters 1 and 2, the operational core of the corpus, are positive for selection and negative for volume. Because the first model is estimated on answers within threads and the other two on questions, this describes distinct association profiles, not a trade-off any single post faces.

## 5 Discussion

### 5.1 Two association profiles, measured on different units

These analyses map, at two levels of granularity, which content co-occurs with two structural outcomes. At the word level, concrete operational vocabulary — "distribution," "warehouse," "center," "base" — is associated with an answer being selected over its competitors, while generic or deferential vocabulary is associated with being passed over (§ 4.1); broad civic and infrastructure vocabulary marks high-engagement questions, while narrow transactional vocabulary marks questions that stop after a single response (§ 4.2). The same contrast reappears at the cluster level as two distinct association profiles (§ 4.4).

What this design does not show bears emphasizing. Best-answer selection is measured on answer texts within threads, engagement on question texts across threads. A cluster positive on one outcome and negative on the other is therefore not evidence that a single post trades one for the other: the two coefficients describe different documents, written by different users for different purposes. The recurrence of operational vocabulary on the selected-answer side and the low-engagement-question side is better read as a property of the question types a keyword-defined corpus contains — narrow, decisively answerable queries versus broad, debatable ones. Testing a genuine trade-off would require both outcomes defined on the same unit, for example a thread-level joint model of answer count and best-answer designation.

This is consistent with work on content-context congruence in Q&A, where answers matching the question's language attributes are judged more helpful [26] and thematic consistency predicts helpfulness while sentiment consistency does not [16], and with evidence that question framing shapes the breadth of engagement [8]. Because the asker chooses the best answer, the outcome reflects asker satisfaction rather than verified correctness [29].

### 5.2 An exploratory typology and its limitations

The cluster-level patterns of § 4.4 are grouped by the sign and significance of coefficients from separate models. No theory predicted these groupings, and the clusters were discovered from the corpus rather than defined a priori, so the grouping is an exploratory typology that generates hypotheses — for instance, that answerability and discussability are distinct content dimensions rather than opposite ends of one axis. Two clusters (calendar/dates, job-seeking-and-Q&A-framing) are lexical or meta-discursive artifacts and are excluded from substantive interpretation.

Five further limitations bound what the results can support. First, the pipeline runs on machine translation. Review of a stratified 50-post sample (§ 3.2) found core topical nouns preserved and losses concentrated in proper nouns and rare loanwords, so error acts mainly as attenuation; systematic differences across term classes cannot be ruled out, and replication with a Japanese morphological pipeline remains the strongest check.

Second, the analysis is restricted to nouns, so it characterizes topical content rather than wording in the broader sense; verbs, adjectives, politeness markers, sentiment, and discourse structure are excluded by design and may carry substantial signal.

Third, the clusters group word vectors, not documents, and are one reproducible partition of one embedding rather than a claim that twelve is the true number of semantic groups. The underlying vocabulary neighbourhoods survive retraining without reply and thanks posts (Table 10), but the partition is sensitive to the PCA dimensionality and to re-clustering. Roughly half of noun-token frequency remains unclassified background.

Fourth, several confounders go unmeasured. Answerer reputation and history, asker characteristics, question category, thread exposure, and formatting features are absent from the scraped records, and because usernames were deliberately not retained (Declarations), repeat-poster dependence cannot be modeled. All results are conditional associations.

Fifth, the corpus covers a single keyword on a single Japanese platform over 2004–2026, so the findings describe logistics-related discourse on Yahoo! Chiebukuro and should not be assumed to transfer without replication.

### 5.3 Future research

Three directions follow. A thread-level joint model of answer volume and best-answer designation would test on a single unit the trade-off this design can only describe as two profiles. Extending the feature set beyond nouns — sentiment, register, politeness, question type, and concreteness norms applied to the original Japanese — would connect these lexical associations to the pragmatic and social dimensions of cQA participation. Finally, pairing yearly cluster prevalence with a chronology of logistics-affecting events would address whether external events reshape which content attracts engagement, a question the present cross-sectional design cannot answer.

## 6 Conclusion

In this exploratory study, we investigated what types of content are associated with two structural outcomes in logistics-related discussions on Yahoo! Chiebukuro: best-answer selection and sustained engagement. By examining both individual nouns and semantic clusters across a keyword-defined corpus, we show that these outcomes have distinct content profiles rather than reflecting a single dimension of content value. Methodologically, we modeled best-answer selection within question threads and decomposed engagement into two stages: whether a question receives any response and, conditional on receiving one, how many responses it attracts. This design separates answerability from subsequent discussion volume. Controlling for answer length and question-answer noun overlap was also important, as both are strongly associated with best-answer selection and could otherwise obscure content-level associations. Substantively, the results indicate that content associated with successfully selected answers generally differs from content associated with questions that attract responses. Thus, "answerability" and "discussability" appear not to lie at opposite ends of a single axis but to represent distinct dimensions of content-outcome association. These findings are conditional associations observed within a single keyword-defined corpus on a single platform, rather than causal relationships. Future research could extend this framework by incorporating social-linguistic features such as sentiment, writing style, and register, and by linking engagement patterns to external logistics-related events through time-series analysis. Such extensions could clarify whether the observed content-outcome profiles generalize beyond topical nouns and how external events shape engagement. Finally, testing a true trade-off between answerability and discussability would require a joint model of response volume and best-answer designation at a common unit of analysis.

## 7 Appendix

For each of the twelve HDBSCAN-detected clusters, the contribution is that cluster's share of total noun-token frequency among the 800 candidate nouns; clusters are ordered by contribution, and the cumulative column sums down the table. The final row pools the nouns HDBSCAN left unassigned.

See Tables 2 and 3 here.

**Table 2** Cluster Contribution and Cumulative Contribution

| Cluster | Short name | Words (*n*) | Frequency | Contribution | Cumulative |
|---|---|---|---|---|---|
| 1 | Work and employment | 40 | 241,414 | 14.01% | 14.01% |
| 2 | Delivery and ordering | 38 | 117,784 | 6.84% | 20.85% |
| 3 | Places and regions | 43 | 93,683 | 5.44% | 26.28% |
| 4 | Nations and geopolitics | 25 | 75,466 | 4.38% | 30.66% |
| 5 | Job seeking and Q&A framing | 19 | 69,437 | 4.03% | 34.69% |
| 6 | Road and rail transport | 24 | 66,124 | 3.84% | 38.53% |
| 7 | Cost, tax and government | 29 | 62,063 | 3.60% | 42.13% |
| 8 | Opinion and interpretation | 22 | 44,795 | 2.60% | 44.73% |
| 9 | Education | 16 | 32,482 | 1.89% | 46.62% |
| 10 | Family and workplace relations | 20 | 28,515 | 1.65% | 48.27% |
| 11 | Infrastructure and strategy | 10 | 17,732 | 1.03% | 49.30% |
| 12 | Calendar and dates | 16 | 16,318 | 0.95% | 50.25% |
| — | (unclassified/noise) | 498 | 857,166 | 49.75% | 100.00% |

The twelve clusters jointly account for 50.3% of noun-token frequency among the 800 candidate nouns; the remainder (49.7%) does not form a sufficiently dense neighborhood under HDBSCAN and is treated as unclassified background rather than forced into a cluster (§ 3.5). The largest cluster holds 14.0% of noun-token frequency and 40 of the 302 classified nouns (13.2%), satisfying the ≤ 50% balance constraint of the selection rule

**Table 3** Parameters and Software

| Stage | Settings |
|---|---|
| Software | Python 3.9; gensim 4.4.0, scikit-learn 1.6, statsmodels 0.14, NLTK 3.9.2, CTranslate2 4.8.1, SentencePiece, umap-learn 0.5, pandas 2.x, numpy 1.26 |
| Translation | Argos Translate (argos_translate_lt 1.12.1), model package translate-ja_en v1.1 (OPUS-MT ja→en converted to CTranslate2), SentencePiece subword tokenizer; sentence-boundary chunking (≤ 200 characters/chunk, no truncation); beam size 2, max 512 decoding tokens; 10 inter-op/1 intra-op threads, batches capped at 1,024 tokens |
| Noun extraction | NLTK POS tagging, tags NN* (NN, NNS, NNP, NNPS), WordNet-lemmatized, lowercased; alphabetic tokens of ≥ 3 characters or numeric tokens of ≥ 4 digits; English stopword list plus a hand-added list of generic and translation-artifact tokens (thing, way, case, please, thank, new, contraction fragments don/isn/doesn…, and the OCR-like artifact gnation); full list distributed with the code |
| Noun vocabulary | 6,792 distinct nouns at min_count = 15; the 800 most frequent are the clustering candidates |
| Word2Vec | CBOW (sg = 0), 100 dimensions, window 5, min_count = 15, 10 epochs, seed 42, single-threaded (workers = 1) — required for bit-reproducible vectors and clusters |
| Clustering input | L2-normalize the 800 noun vectors, then PCA to 20 dimensions |
| HDBSCAN parameters | min_cluster_size = 10, min_samples = 1, cluster_selection_epsilon = 0.0, Euclidean metric on the PCA space |
| Selection rule | Grid over PCA dims {10,12,15,18,20} × min cluster size {6,8,…,20} × min samples {1,2,3} × epsilon {0.0,0.2,0.3,0.4,0.5}; admissible solutions have 6–15 clusters and a largest cluster ≤ 50% of classified nouns; among these, maximize frequency-weighted coverage |
| Clustering result | 12 clusters; 302/800 nouns classified (37.8%); 50.3% frequency-weighted coverage (Table 2); largest cluster 40 nouns (13.2%) |
| UMAP | Cosine metric, 15 neighbors, minimum distance 0.15, seed 42, single-threaded (n_jobs = 1) |
| Keyness | Monroe et al., [20] informative-Dirichlet-prior log-odds z-scores; 500 within-thread label permutations preserving thread sizes; threshold robustness 75/25 vs. 66/33 split |
| Regressions | Best answer: conditional logit stratified by question ID. Question models: logit for any-answer on all questions, NB2 for engagement among answered questions. Benjamini–Hochberg FDR applied within each model across the twelve cluster terms |

### 7.1 Regression analysis table

Benjamini–Hochberg FDR correction is applied within each model across the twelve cluster terms only; control terms are not part of the correction family and their p_FDR is left blank Tables 4, 5, 6 and 7.

**Table 4** Best-answer selection — conditional logit stratified by question

| Cluster | β | SE | 95% CI (β) | OR | 95% CI (OR) | *p* | p_FDR |
|---|---|---|---|---|---|---|---|
| 1 | 0.247 | 0.031 | [0.185, 0.308] | 1.280 | [1.203, 1.361] | < 0.001 | < 0.001 |
| 2 | 0.234 | 0.030 | [0.175, 0.293] | 1.264 | [1.191, 1.341] | < 0.001 | < 0.001 |
| 3 | 0.200 | 0.037 | [0.127, 0.273] | 1.222 | [1.136, 1.314] | < 0.001 | < 0.001 |
| 4 | 0.275 | 0.037 | [0.203, 0.347] | 1.317 | [1.226, 1.415] | < 0.001 | < 0.001 |
| 5 | −0.088 | 0.035 | [−0.158, −0.019] | 0.915 | [0.854, 0.981] | 0.013 | 0.014 |
| 6 | 0.230 | 0.038 | [0.155, 0.305] | 1.258 | [1.167, 1.356] | < 0.001 | < 0.001 |
| 7 | 0.101 | 0.034 | [0.034, 0.169] | 1.106 | [1.034, 1.184] | 0.003 | 0.006 |
| 8 | −0.047 | 0.031 | [−0.108, 0.014] | 0.954 | [0.898, 1.014] | 0.128 | 0.128 |
| 9 | −0.136 | 0.049 | [−0.232, −0.040] | 0.873 | [0.793, 0.961] | 0.006 | 0.007 |
| 10 | −0.117 | 0.043 | [−0.200, −0.034] | 0.889 | [0.818, 0.967] | 0.006 | 0.007 |
| 11 | 0.296 | 0.043 | [0.211, 0.381] | 1.344 | [1.235, 1.463] | < 0.001 | < 0.001 |
| 12 | 0.134 | 0.046 | [0.043, 0.225] | 1.144 | [1.044, 1.253] | 0.004 | 0.006 |
| log(length) | 0.953 | 0.022 | [0.909, 0.996] | 2.593 | [2.482, 2.708] | < 0.001 | — |
| Within-thread order | 0.131 | 0.010 | [0.111, 0.151] | 1.140 | [1.118, 1.163] | < 0.001 | — |
| Question–answer noun overlap | −2.070 | 0.178 | [−2.418, −1.722] | 0.126 | [0.089, 0.179] | < 0.001 | — |

Answer-level; $n = 51{,}084$ answers in 15,104 threads with a designated best answer. Effect size is the odds ratio

**Table 5** Receives any answer — logistic regression, all questions

| Cluster | β | SE | 95% CI (β) | OR | 95% CI (OR) | *p* | p_FDR |
|---|---|---|---|---|---|---|---|
| 1 | −0.025 | 0.031 | [−0.086, 0.036] | 0.975 | [0.917, 1.036] | 0.417 | 0.455 |
| 2 | −0.033 | 0.029 | [−0.089, 0.024] | 0.968 | [0.914, 1.024] | 0.254 | 0.382 |
| 3 | −0.394 | 0.031 | [−0.454, −0.334] | 0.674 | [0.635, 0.716] | < 0.001 | < 0.001 |
| 4 | −0.152 | 0.034 | [−0.219, −0.085] | 0.859 | [0.803, 0.918] | < 0.001 | < 0.001 |
| 5 | −0.211 | 0.031 | [−0.272, −0.150] | 0.809 | [0.762, 0.860] | < 0.001 | < 0.001 |
| 6 | 0.040 | 0.035 | [−0.029, 0.108] | 1.041 | [0.972, 1.114] | 0.255 | 0.382 |
| 7 | 0.153 | 0.038 | [0.080, 0.227] | 1.166 | [1.083, 1.255] | < 0.001 | < 0.001 |
| 8 | 0.033 | 0.035 | [−0.037, 0.102] | 1.033 | [0.964, 1.108] | 0.353 | 0.455 |
| 9 | −0.005 | 0.040 | [−0.083, 0.072] | 0.995 | [0.920, 1.075] | 0.894 | 0.894 |
| 10 | −0.087 | 0.039 | [−0.164, −0.010] | 0.917 | [0.849, 0.990] | 0.027 | 0.055 |
| 11 | −0.044 | 0.051 | [−0.145, 0.057] | 0.957 | [0.865, 1.059] | 0.394 | 0.455 |
| 12 | −0.114 | 0.044 | [−0.200, −0.029] | 0.892 | [0.819, 0.972] | 0.009 | 0.021 |
| log(length) | −0.037 | 0.021 | [−0.079, 0.004] | 0.963 | [0.924, 1.004] | 0.075 | — |
| Posting year (centred on 2015) | 0.325 | 0.003 | [0.319, 0.331] | 1.384 | [1.375, 1.393] | < 0.001 | — |

Question-level; $n = 40{,}836$ questions (35 dropped for a missing timestamp or empty text), of which 24,756 received at least one response. Effect size is the odds ratio

### 7.2 Cluster validation

Every cluster's coherence exceeds its random baseline by a factor of at least six, so the partition is not an arbitrary slicing of a homogeneous space. Closeness to "logistics" orders the clusters much as their labels suggest: delivery and ordering is nearest the domain keyword, work and employment and infrastructure and strategy follow, and the clusters concerned with opinion, family, and education sit at or below zero Tables 8, 9 and 10.

The two parameter-free tests indicate that reply and thanks text does not drive the semantic structure. Test (c) does not reproduce the reported partition, but alongside the PCA-dimensionality result in A4b it is better read as instability in HDBSCAN's discrete cluster extraction than as a property of the corpus Table 11.

**Table 6** Number of responses once answered — negative binomial (NB2)

| Term | β | SE | 95% CI (β) | IRR | 95% CI (IRR) | *p* | p_FDR |
|---|---|---|---|---|---|---|---|
| 1 | −0.111 | 0.011 | [−0.132, −0.090] | 0.895 | [0.877, 0.914] | < 0.001 | < 0.001 |
| 2 | −0.081 | 0.010 | [−0.101, −0.062] | 0.922 | [0.904, 0.940] | < 0.001 | < 0.001 |
| 3 | 0.013 | 0.011 | [−0.009, 0.034] | 1.013 | [0.991, 1.035] | 0.251 | 0.274 |
| 4 | 0.091 | 0.012 | [0.068, 0.114] | 1.095 | [1.070, 1.120] | < 0.001 | < 0.001 |
| 5 | −0.070 | 0.011 | [−0.092, −0.049] | 0.932 | [0.912, 0.952] | < 0.001 | < 0.001 |
| 6 | 0.186 | 0.012 | [0.163, 0.209] | 1.204 | [1.177, 1.232] | < 0.001 | < 0.001 |
| 7 | 0.127 | 0.012 | [0.103, 0.152] | 1.136 | [1.108, 1.164] | < 0.001 | < 0.001 |
| 8 | 0.060 | 0.012 | [0.036, 0.085] | 1.062 | [1.037, 1.088] | < 0.001 | < 0.001 |
| 9 | 0.034 | 0.014 | [0.007, 0.061] | 1.035 | [1.007, 1.063] | 0.013 | 0.017 |
| 10 | 0.108 | 0.014 | [0.081, 0.135] | 1.114 | [1.084, 1.144] | < 0.001 | < 0.001 |
| 11 | −0.001 | 0.018 | [−0.036, 0.034] | 0.999 | [0.964, 1.035] | 0.956 | 0.956 |
| 12 | −0.037 | 0.016 | [−0.067, −0.007] | 0.964 | [0.935, 0.993] | 0.017 | 0.020 |
| Dispersion α (NB2) | 0.132 | 0.004 | [0.124, 0.140] | — | — | < 0.001 | — |

Question-level; $n = 24{,}756$ answered questions. Estimated dispersion $\alpha = 0.132$, confirming overdispersion relative to Poisson. Effect size is the incidence-rate ratio

**Table 7** Best-answer selection — pooled logit with question-clustered standard errors

| Term | β | SE | 95% CI (β) | OR | 95% CI (OR) | *p* | p_FDR |
|---|---|---|---|---|---|---|---|
| 1 | 0.382 | 0.023 | [0.337, 0.426] | 1.465 | [1.401, 1.532] | < 0.001 | < 0.001 |
| 2 | 0.369 | 0.022 | [0.326, 0.412] | 1.446 | [1.386, 1.510] | < 0.001 | < 0.001 |
| 3 | 0.085 | 0.024 | [0.037, 0.132] | 1.089 | [1.038, 1.141] | < 0.001 | < 0.001 |
| 4 | 0.015 | 0.025 | [−0.034, 0.064] | 1.015 | [0.966, 1.066] | 0.547 | 0.547 |
| 5 | −0.040 | 0.026 | [−0.092, 0.012] | 0.961 | [0.912, 1.012] | 0.131 | 0.143 |
| 6 | 0.047 | 0.025 | [−0.001, 0.095] | 1.048 | [0.999, 1.100] | 0.054 | 0.071 |
| 7 | −0.048 | 0.025 | [−0.098, 0.002] | 0.953 | [0.907, 1.002] | 0.059 | 0.071 |
| 8 | −0.103 | 0.025 | [−0.153, −0.054] | 0.902 | [0.858, 0.948] | < 0.001 | < 0.001 |
| 9 | −0.143 | 0.033 | [−0.207, −0.078] | 0.867 | [0.813, 0.925] | < 0.001 | < 0.001 |
| 10 | −0.105 | 0.032 | [−0.167, −0.043] | 0.900 | [0.846, 0.958] | < 0.001 | 0.001 |
| 11 | 0.239 | 0.034 | [0.173, 0.305] | 1.270 | [1.189, 1.356] | < 0.001 | < 0.001 |
| 12 | 0.211 | 0.036 | [0.140, 0.283] | 1.236 | [1.150, 1.327] | < 0.001 | < 0.001 |
| log(length) | 0.647 | 0.017 | [0.613, 0.680] | 1.909 | [1.847, 1.973] | < 0.001 | — |

The specification used in the previous version of this manuscript, retained for comparison. Signs agree with the conditional logit on 11 of 12 clusters; the exception is Cluster 7, positive under stratification and negative when pooled

**Table 8** Coherence and closeness to the domain keyword

| Cluster | Words | Coherence | Random baseline | Ratio | Closeness to "logistics" |
|---|---|---|---|---|---|
| 1 | 40 | 0.302 | 0.029 | 10.4× | + 0.338 |
| 2 | 38 | 0.182 | 0.029 | 6.3× | + 0.519 |
| 3 | 43 | 0.257 | 0.029 | 8.9× | + 0.111 |
| 4 | 25 | 0.265 | 0.029 | 9.1× | + 0.026 |
| 5 | 19 | 0.309 | 0.028 | 11.0× | + 0.182 |
| 6 | 24 | 0.318 | 0.029 | 11.0× | + 0.150 |
| 7 | 29 | 0.221 | 0.030 | 7.4× | + 0.005 |
| 8 | 22 | 0.358 | 0.029 | 12.3× | −0.156 |
| 9 | 16 | 0.385 | 0.028 | 13.8× | −0.045 |
| 10 | 20 | 0.374 | 0.031 | 12.1× | −0.090 |
| 11 | 10 | 0.324 | 0.028 | 11.6× | + 0.308 |
| 12 | 16 | 0.578 | 0.029 | 19.9× | + 0.020 |

Coherence is the mean pairwise cosine similarity among a cluster's members in the Word2Vec space. The random baseline is the mean coherence of 200 random same-size noun sets drawn from the same 800 candidates. Closeness is the cosine similarity between the cluster centroid and the vector for "logistics," the corpus's search keyword

**Table 9** Stability across neighboring parameter settings

| Dims | Min cluster | Min samples | Epsilon | Clusters | ARI vs. reported | Shared classified words |
|---|---|---|---|---|---|---|
| 18 | 10 | 1 | 0.0 | 3 | 0.029 | 301 |
| 25 | 10 | 1 | 0.0 | 10 | 0.552 | 230 |
| 20 | 8 | 1 | 0.0 | 14 | 0.971 | 284 |
| 20 | 12 | 1 | 0.0 | 11 | 1.000 | 292 |
| 20 | 14 | 1 | 0.0 | 11 | 1.000 | 292 |
| 20 | 10 | 2 | 0.0 | 12 | 1.000 | 302 |
| 20 | 10 | 3 | 0.0 | 11 | 0.972 | 224 |
| 20 | 10 | 1 | 0.2 | 12 | 1.000 | 302 |

Mean ARI across the eight settings is 0.82. Six settings reproduce the reported partition almost exactly (ARI ≥ 0.97). The two low values both come from changing the PCA dimensionality rather than an HDBSCAN parameter: at 18 dimensions the density structure collapses into three coarse clusters, and at 25 it fragments into ten only partly aligned ones

Adjusted Rand index (ARI) between the reported solution (PCA 20, minimum cluster size 10, minimum samples 1, epsilon 0.0) and solutions at neighboring settings, computed on the nouns classified in both solutions

**Table 10** Sensitivity to the training corpus

| Test | Tests whether | Params | Result | Baseline | Reading |
|---|---|---|---|---|---|
| (a) Neighbour preservation | top-10 neighbours persist | none | mean 0.852, median 0.900 | 0.026 (random pairing) | neighbourhoods hold |
| (b) Coherence, recomputed | clusters stay internally tight | none | 0.176–0.575, all above baseline | 6.0×–19.9× random sets | clusters hold |
| (c) Re-clustering with HDBSCAN | the partition is recovered | size 10 (also 9, 11) | 2 clusters, ARI 0.014 | reported 12 clusters | not recovered (A4b) |

Word2Vec was retrained on questions and answers only, dropping reply and thanks posts. These posts are numerous but short: they account for 3.3% of noun tokens, so min_count needed no rescaling (the scaled value rounds back to 15). Three tests were run. Tests (a) and (b) answer the substantive question — whether reply and thanks text drives the semantic structure — and neither depends on any clustering parameter; test (c) re-runs the clustering step and therefore does

**Table 11** Complete Cluster Membership

| Cluster | Words (*n*) | Members (descending frequency) |
|---|---|---|
| 1 | 40 | Company, work, year, day, business, office, employee, month, management, hour, department, labor, week, staff, salary, income, holiday, worker, manager, dispatch, morning, night, charge, minute, half, sunday, break, saturday, treatment, overtime, rest, vacation, evening, weekday, shift, working, bonus, allowance, dispatcher, personnel |
| 2 | 38 | Product, delivery, order, reason, shipping, post, distribution, mail, change, amazon, course, luggage, point, condition, result, yamato, custom, date, stock, express, sagawa, flight, purchase, flow, destination, package, inventory, entry, chain, arrival, status, view, courier, shipment, step, sorry, clearance, baggage |
| 3 | 43 | City, area, tokyo, prefecture, osaka, region, sea, earthquake, port, base, hokkaido, east, damage, river, edo, kanto, north, island, kyushu, kansai, kyoto, tohoku, west, fukushima, south, fukuoka, chiba, nagoya, okinawa, great, saitama, sapporo, district, bay, pacific, sendai, aichi, hiroshima, niigata, shikoku, hokuriku, tsunami, kanagawa |
| 4 | 25 | Japan, country, power, China, state, war, economy, united, Korea, army, force, energy, Russia, movement, Taiwan, Korean, peninsula, asia, Kong, Hong, Germany, defense, Chinese, Ukraine, continent |
| 5 | 19 | Job, question, understanding, experience, interview, answer, skill, first, knowledge, advice, reply, career, hello, sentence, recruitment, note, comment, engineering, engineer |
| 6 | 24 | Car, road, line, station, route, traffic, railway, expressway, train, bus, speed, shinkansen, distance, highway, passenger, congestion, ferry, tunnel, jam, taxi, bicycle, parking, main, bike |
| 7 | 29 | Cost, tax, government, amount, party, rate, increase, debt, consumption, member, ministry, reduction, standard, rise, investment, regulation, inflation, official, minister, asset, fund, administration, bond, prime, election, governor, declaration, decrease, democratic |
| 8 | 22 | Person, word, story, feeling, idea, opinion, sense, questioner, others, meaning, effort, thought, eye, impression, face, heart, well, mouth, voice, personality, let, look |
| 9 | 16 | School, university, student, test, class, english, language, study, science, teacher, examination, certificate, grade, faculty, subject, graduation |
| 10 | 20 | Woman, family, parent, friend, workplace, head, man, bos, mother, husband, father, supervisor, men, wife, senior, attitude, brother, son, harassment, colleague |
| 11 | 10 | Support, development, resource, security, infrastructure, activity, network, growth, strategy, expansion |
| 12 | 16 | End, middle, march, beginning, april, june, december, january, february, may, july, october, september, november, august, start |

All 302 classified nouns, listed by cluster in descending token frequency. The 498 unclassified nouns are omitted (they are treated as background; § 3.5)

**Author contributions**
H.N. wrote the main manuscript text, prepared all figures and reviewed the manuscript.

**Funding**
This research did not receive any specific grant from funding agencies in the public, commercial, or not-for-profit sectors.

**Data availability**
No datasets were generated or analysed during the current study.

## Declarations

**Ethics approval and consent to participate**
This study used publicly available, anonymized online content and did not involve direct interaction with human participants. Therefore, ethical approval was not required.

**Consent for publication**
Not applicable.

**Competing interests**
The authors declare no competing interests.

**Research ethics and data statement**
All analyzed content consists of question-and-answer posts publicly viewable on Yahoo! Chiebukuro at the time of collection; no private messages, account credentials, or other non-public data were accessed. The scraped records retain only post text, post/question identifiers, timestamps, and structural tags (question/best answer/other answer/reply/thanks); no usernames, poster IDs, or other author-identifying fields were collected or retained. All reported results are aggregate statistics (word/cluster frequencies, z-scores, regression coefficients) over thousands of posts; no individual post is reproduced or attributed to an identifiable author in this paper.

## Publisher's note

Springer Nature remains neutral with regard to jurisdictional claims in published maps and institutional affiliations.